# Machine Learning-Assisted E-jet Printing of Organic Flexible Biosensors

Mehran Abbasi Shirsavar [a], Mehrnoosh Taghavimehr [a], Lionel J. Ouedraogo [a], Mojan Javaheripi [b], Nicole N. Hashemi [a, c], Farinaz Koushanfar [b], Reza Montazami [a*]

[a] Department of Mechanical Engineering, Iowa State University, Ames, IA 50011, USA

[b] Department of Electrical and Computer Engineering, University of California, San Diego, CA 92093, USA

[c] Department of Mechanical Engineering, Stanford University, Stanford, CA 94305, USA

[*] reza@iastate.edu

**Abstract:**

Electrohydrodynamic-jet (e-jet) printing technique enables the high-resolution printing of complex soft electronic devices. As such, it has an unmatched potential for becoming the conventional technique for printing soft electronic devices. In this study, the electrical conductivity of the e-jet printed circuits was studied as a function of key printing parameters (nozzle speed, ink flow rate, and voltage). The collected experimental dataset was then used to train a machine learning algorithm to establish models capable of predicting the characteristics of the printed circuits in real-time. Precision parameters were compared to evaluate the supervised classification models. Since decision tree methods could not increase the accuracy higher than 71%, more advanced algorithms are performed on our dataset to improve the precision of model. According to F-measure values, the K-NN model (k=10) and random forest are the best methods to classify the conductivity of electrodes. The highest accuracy of AdaBoost ensemble learning has resulted in the range of 10-15 trees (87%).

**Keywords:** Machine learning, flexible electronics, sensors, e-jet printing,

# 1. Introduction

Inkjet printing has been used as an additive manufacturing technique by which a given ink material (typically functional) is deposited in layers on a substrate (typically inert) to obtain patterns. Inkjet printing is an accurate, reliable, and cost-effective fabrication method [2]. However, to eliminate the need for a pulsating pressure to deposit ink onto the substrate, increase control over the droplets' size and shape, and obtain a higher resolution, the electrohydrodynamic-jet (e-jet) printing method was introduced as a modification method of ink jet printing. This class of manufacturing has a wide range of applications [3-6], with many emerging in the biomedical field especially in fabrication of three-dimensional fibrous scaffolds (cartilage [7], tendon [8] and blood vessels) [1].

E-jet printing utilizes a high-voltage difference between an electrically conductive nozzle and a substrate to generate an electric field that draws the ink from the nozzle [9], which can be used to create high-resolution prints [10]. In addition to the electric field, there are other key print parameters that significantly affect the formation of ink droplets, including flow rate, standoff distance, nozzle-substrate relative velocity, and the type of ink materials, to name few examples [11]. These parameters must be controlled, and optimized, to obtain and maintain a consistent droplet ejection and deposition on the substrate.

Soft electronics are introduced as the next generation electronics with the ability to keep their electrical conductivity under mechanical loadings [12-16]. The use of bioelectronics as healthcare devices provide the opportunity to understand the biological signals. The biocompatibility organic electronics has made them a potential tool at the interface of biological systems [17-20]. Among different types of organic conductors, graphene ink has been of great interest because of several beneficial properties, including high surface area, high thermal conductivity, and biocompatibility as a crucial feature to interact with biomarkers.

The graphene ink was then used to fabricate high resolution sensors that can detect the alteration of biological systems through the analysis of electrical responses.
The major challenge in fabricating graphene sensors using e-jet printing is controlling the processing parameters that can affect the conductivity of the device. Training machine learning (ML) methods based on the manufacturing data and employing the resultant algorithms to identify and apply optimum manufacturing parameters would enable high-fidelity e-jet manufacturing of soft bioelectronics [21, 22]. Such algorithms can predict optimum manufacturing conditions based on the previously established correlations among parameters [23, 24]. As such, the development of these algorithms has been the focus of several studies [25, 26].

In one study Zhao *et al.* designed a regression-based model to find a connection between chemical composition of Cu alloys and the resulted physical properties including hardness and conductivity that could develop better hardness and electrical conductivity compared to the experimental results [28]. In another study by Shi *et al.* multilayer perceptron method was used to classify several parameters including bio-ink viscosity and nozzle diameter of their drop-on-demand cell printing system to predict the likelihood of satellite droplets being printed [29]. These examples illustrate the wide range of applications for ML and the potential it possesses to improve the accuracy and efficiency of data processing. In 2019 Shi *et al.* utilized two ML algorithms to optimize printing parameters for drop-on-demand piezoelectric printing using bio-ink. Their work involved identifying an optimal range for the viscosity and surface tension of the bio-ink as well as the voltage applied between nozzle and substrate [30]. These factors must now be balanced with other properties that affect the suitability of the materials as a sensor. Wu *et al.* utilized ensemble learning with four different types of algorithms (including random forests, LASSO, XGBoost and SVR) to predict droplet velocity and volume of ink jet process. The performance of models was evaluate through root-mean-square error (RMSE), relative error (RE), and coefficient of determination ($R^2$). The results showed the sufficient accuracy of the models in predicting the experimental results.[31].

In this study, we have employed the supervised learning method due to its simplicity and relatively straightforward classification system to predict the properties of e-jet printed graphene-based soft biosensors by evaluating the applied manufacturing parameters.

The supervised learning methods were trained based on the experimental data set. The accuracy of the applied algorithms to predict the conductivity of electrodes will be evaluated. Then, the most accurate model can be used to estimate the electrical properties of a sensor as a result of changing the processing parameters, which can save time, ink, and guarantee the functionality of the sensor before the production process.

## 2. Materials and Methods

Few-layer graphene (FLG) ink was synthesized from graphite (Sigma-Aldrich 7782-42-5) using a method we developed earlier [6]. Briefly, an in-house-made vibro-energy mill ball-milling exfoliation technique was coupled with sheer force to obtain high yields of FLG and Bovine Serum Albumin (BSA) protein (Sigma-Aldrich 9048-46-8) was used as the stabilizing agent.

Poly (4,4'-oxydiphenylene-pyromellitimide) *(i.e.,* Kapton) film of a thickness of 0.06 mm (Addicore) was used as a flexible substrate. The film surface was chemically treated to increase hydrophilicity. The films were washed and rinsed with DI water and acetone, and then submerged in an aqueous solution of poly sodium 4-styrenesulfonic (PSS) (Aldrich 25704-18-1) (12 mg mL$^{-1}$) mixed with NaCl (0.5 mg mL$^{-1}$) for 20 minutes; then, submerged in a solution of polyethyleneimine (PEI) (Sigma-Aldrich 9002-98-6) in DI water (30 mg mL$^{-1}$) and NaCl (0.5 mg mL$^{-1}$) for another 20 minutes. Lastly, the film was rinsed with DI water and dried at room temperature.

Platform of a Prusa i3 MK3S 3D printer was modified to an e-jet printer and was coupled with a syringe pump (Genie Touch, serial number GT0589) and high-voltage power source to print FLG ink on the substrates. Print parameters including stand-off distance, applied potential difference, flow rate of the ink, and nozzle velocity were controlled using a CNC G-code controller.

The production of graphene was evaluated using PerkinElmer Lamda25 UV-visible spectroscopy. To evaluate the performance of the electrodes, the electrical properties was measured by changing the voltage over the electrode. Scanning electron microscopy (SEM) images were acquired using a JCM-6000 NeoScope Benchtop SEM (JOEL) with an accelerating voltage of 15 kV.

## 3. Results and Discussions

*Production of graphene electrodes*

The production of graphene using direct Liquid Phase Exfoliation (LPE) has been used for inkjet printing process. Since the graphene platelets tend to aggregate or settle in water, the BSA was used as the stabilizing agent in combination with sheer force to ensure the formation of stable water-based graphene [38]. LPE is a cost-effective, organic solvent-free method allowing biocompatible and large-scale production of graphene for electrochemical biosensor platforms [33]. As shown before, the water-based graphene was selected as highly conductive components of biosensors [34] . The homogeneity of the graphene ink affects the quality of the print since the printed particles must be dispersed enough to prevent clogging in the nozzle and create a connected print line after drying the graphene solution [35, 36].

The production of graphene was investigated using UV-vis spectrum. The absorption spectrum showed a strong absorption band at 270 nm, which is the characteristic of graphene [39].

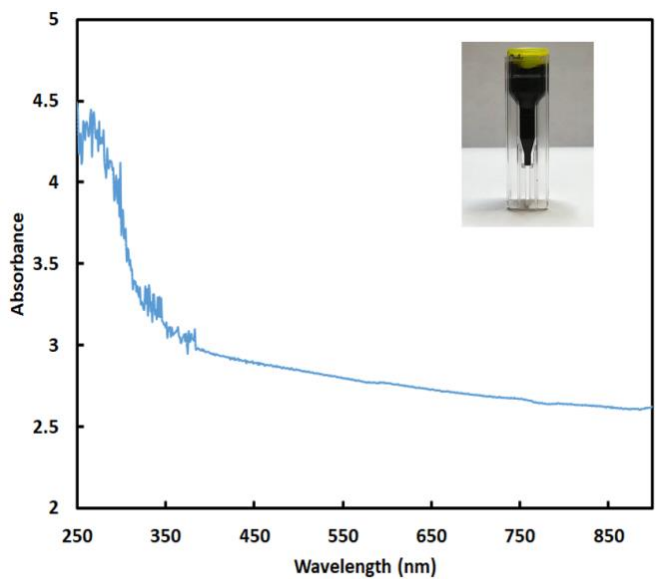

**Figure 1.** UV-visible spectra for synthesized graphene ink.

Presented in **Figure 2** are optical images of a typical interdigitated graphene electrode e-jet printed on a Kapton substrate. The interdigitated electrode pattern was used to increase contact with the specimen. CAD design of organic electrode (Figure S2) shows the high resolution of electrodes were fabricated with the electrode thickness of 350±5 µm and finger spacing of 400±5 µm.

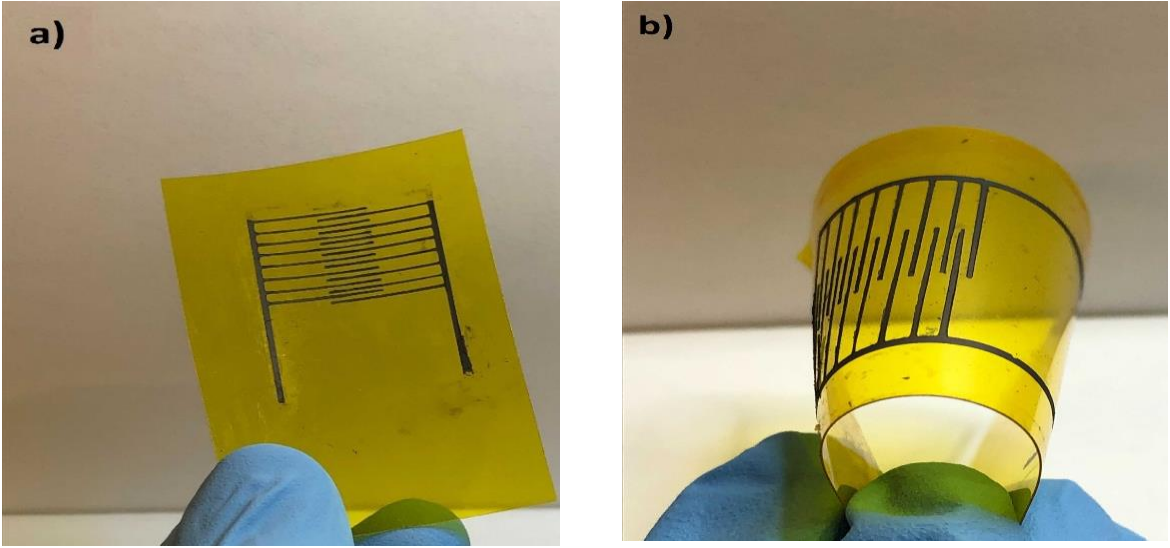

**Figure 2.** a) E-jet printed of organic graphene on flexible substrate b) Flexibility of organic sensors under bending.

*Data collection of E-jet printing process*

The effect of variation of processing parameters on the final property of electrodes can be investigated using supervised classification methods. Monitoring the existing patterns of the existing data can minimize the production of defective electrodes. Several classification models were applied on the acquired experimental data of e-jet manufacturing for the analysis of electrical properties to test the learning characteristics of different ML techniques.

In the first step, the printing parameters were altered for a set of 240 samples, and conductance of the resultant e-jet printed electrode was measured as a means of evaluating the electrodes performance. The conductance of each electrode was deduced from I-V curves obtained between -1 and +1 volts. According to the data in table 1(a), the processing parameters include the nozzle speed, voltage between print head and substrate and the flow rate were changed. The data preprocessing was performed on the dataset by defining the range of resistance to label the output parameter as low-conductance (class=0) and high-conductance (class=1) (table 1b).

**Table 1.** a) The used Print Parameters in fabrication of graphene sensors, b) A sample of collected dataset.

a)

| Print Parameters | | |
|---|---|---|
| Nozzle Speed (mm/min) | Voltage (kV) | Flow rate (μl/min) |
| 300 | 1 | 15 |
| 500 | 2 | 12 |
| 700 | 3 | 10 |
|  | 4 | 9 |
|  |  | 6 |
|  |  | 3 |

b)

| Nozzle Speed (mm/min) | Voltage (kV) | Flowrate (μl/min) | Conductivity |
|---|---|---|---|
| 300 | 0.0 | 16 | 1 |
| 300 | 0.5 | 15 | 1 |
| 300 | 0.0 | 15 | 1 |
| 300 | 1.0 | 16 | 1 |
| 400 | 1.0 | 15 | 1 |

*Characterization of printed samples*

    Scanning Electron Microscopy (SEM) analysis was used to investigate the morphology of the printed electrodes and compare the number of defects on the graphene path of printed lines with a range of resistance. The formation of defects on the printed electrodes implies differences of printing conditions. At nozzle speeds higher than 500 mm/min, measured resistance increased significantly to 241 Ω/sqr, which is classified as low-conductance sample (figure 3d). While at lower speeds, fewer voids are observed on the surface of the printed graphene patterns (figure 3a). By increasing the rate of ink flow, the printed patterns are more cohesive with a continuous path of interconnected graphene flakes that can increase the probability of creating a more conductive graphene line. SEM characterization can be used to connect the visualization characteristics of samples to the final electrical properties.

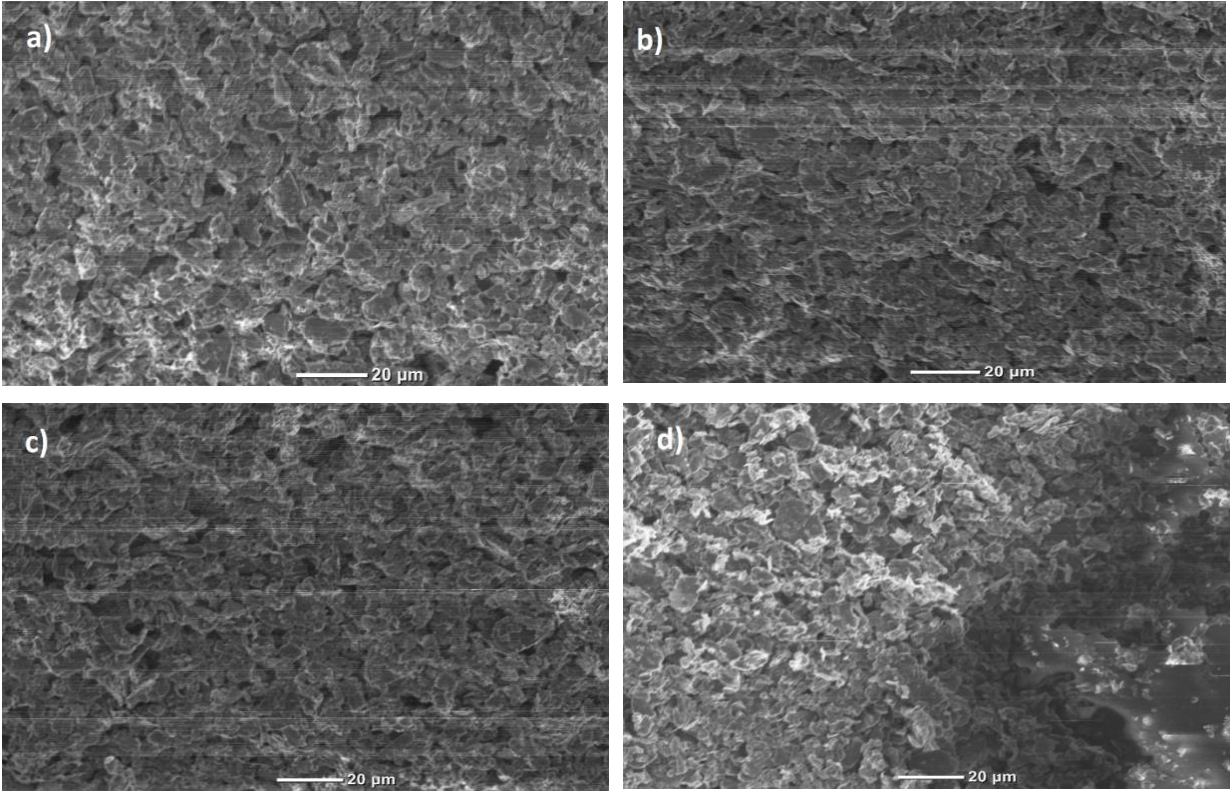

**Figure 3.** SEM images of the surface of E-jet printed graphene electrodes by altering the processing parameters resulted in different measured resistances: (a) Resistance= 35 Ω/sqr, (b) Resistance= 62 Ω/sqr, (c) Resistance= 143 Ω/sqr, and (d) Resistance= 241 Ω/sqr.

*Applying ML models on the data set*

CART (Classification and Regression Trees) package was used to perform the decision tree algorithm using the rpart model that uses a greedy algorithm on the training data to split the tree. The default parameters of CART were used to create the tree shown in figure 4. The results suggested that the nozzle speed has the highest association to the quality of the printed electrodes, as it is calculated as the root node. Furthermore, the feature importance was investigated using random forest measurement by defining a Gaussian classifier model. figure 5 (b) shows the highest importance is related to nozzle speed following by flowrate importance.

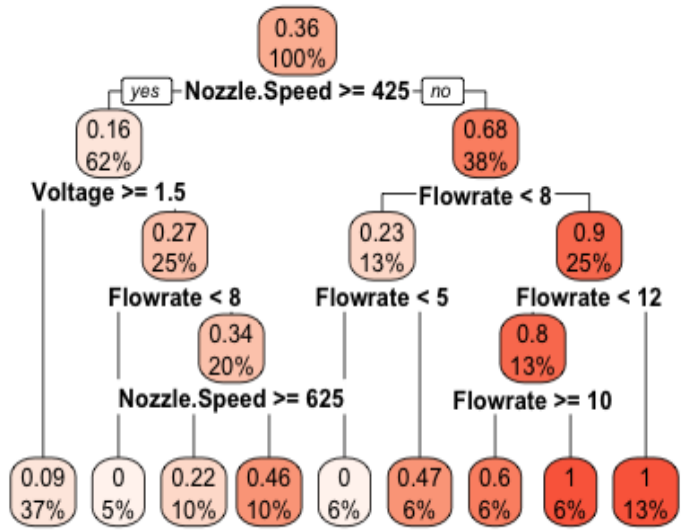

**Figure 4.** The default decision tree chart extracted from CART model.

Cross-Validation (CV) and Bootstrap resampling methods were then used to improve the accuracy of the model. Results for a 10-fold CV, as presented in Figure 5a, exhibited an increase of accuracy to 0.72 at a complexity parameter (cp) of 0.2. Therefore, it is expected to acquire higher accuracies by increasing cp values up to 0.2 in pruned trees.

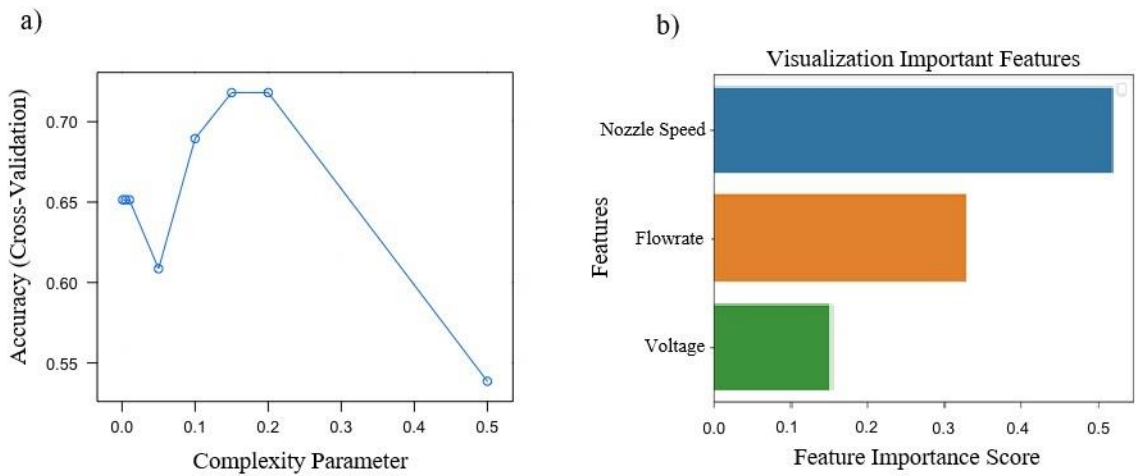

**Figure 5.** a) Accuracy (cross-validation) versus complexity parameter of default decision tree, b) Feature importance investigation using random forest method.

To evaluate the improvement of the performance of the tree, cp was increased to 0.05 to prune the tree that can decrease the cross-validated error. Pruning is implemented to remove the low power splits and reduces overfitting on data. Then, cp was set on 0.2 to obtain the largest accuracy and make a highly pruned tree that eliminates most of the nodes that cannot provide enough information.

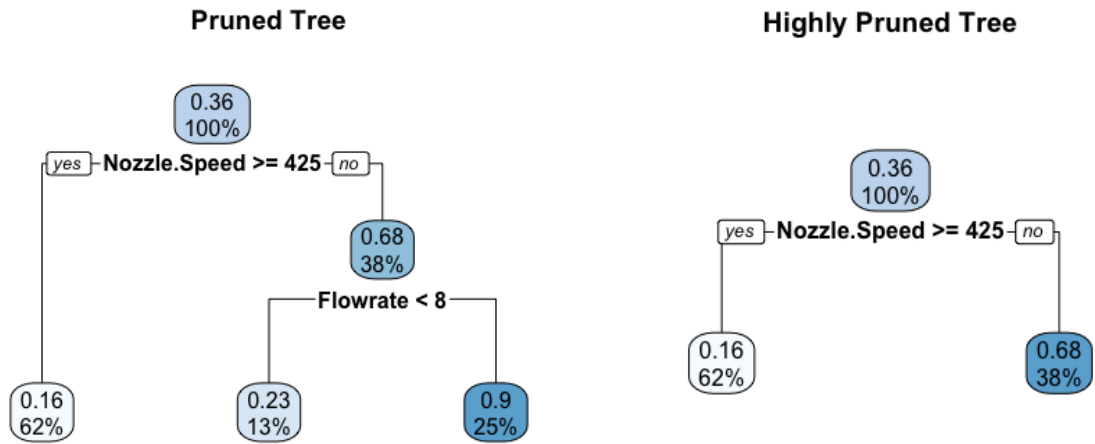

**Figure 6.** Pruning the default decision tree by increasing complexity parameter (cp) to a) Pruned tree, cp=0.05, b) Highly pruned tree, (cp=0.2).

The result of the confusion matrix for trees is provided in table 2. It is very likely for the default tree to overfit the data and cause the poor performance of the model. The lower variance can then be achieved by decreasing the number of splits.

Since accuracy is not the most reliable metric in evaluating a model, we will compare the sensitivity parameters, which is defined as the true positive rate (also known as recall). According to table 2, despite the increase of accuracy by pruning the tree, the sensitivity decreased in highly pruned trees.

**Table 2.** Confusion matrix tables for default tree, pruned tree, and highly pruned tree models.

| | **Default tree** | | | **Pruned tree** | | | **Highly Pruned tree** | |
|---|---|---|---|---|---|---|---|---|
| | sensor.predictions | | | sensor.predictions | | | sensor.predictions | |
| | 0 | 1 | | 0 | 1 | | 0 | 1 |
| 0 | 142 | 12 | 0 | 149 | 5 | 0 | 125 | 29 |
| 1 | 20 | 65 | 1 | 31 | 54 | 1 | 24 | 61 |

Type II error (FN) decreased significantly in the pruned tree; however, it increased again by highly pruning the model. This error can have important consequences on the results and deviate the model from an ideal case.

The random forest can reduce the correlation between tress that improves the variance reduction. This approach was tested on the data as an ensemble model which generates a number of trees, trained on the part of the overall dataset. The second method is the K-Nearest Neighbors algorithm (KNN) as a flexible, non-parametric classification model, where increasing the number of neighbors can create a more stable model. The last supervised model applied is logistic regression.

In order to evaluate and compare the performance of supervised models, different parameters were defined below and calculated accordingly.

$$\text{Accuracy} = \frac{(TP + TN)}{(TP + FP + FN + TN)} \tag{1}$$

$$\text{Mis} - \text{Classification} = (1 - \text{Accuracy}) \tag{2}$$

$$\text{Precision} = \frac{TP}{(FN + TP)} \tag{3}$$

$$\text{Sensitivity (Recall)} = \frac{TP}{(TP + FP)} \tag{4}$$

$$\text{F1 Score} = \frac{2 \times (\text{Precision} \times \text{Sensitivity})}{(\text{Precision} + \text{Sensitivity})} \tag{5}$$

$$\text{Kappa} = \frac{(\text{total accuracy} - \text{rendom accuracy})}{(1 - \text{random accuracy})} \tag{6}$$

The kappa statistics measures how correctly ML classifiers classify the instances. The value of kappa for RF and K-NN model (K=10) is classified as substantial strength; in addition, this statistic is in the moderate range for the K-NN model (K=3) and logistic regression model. F-measure combines the precision and sensitivity and is more preferable compared to accuracy in our dataset, where the two output classes are imbalanced. Based on F-measure, RF and K-NN model (K=10) perform better in predictability.

Table 3. Precision parameters of supervised models.

| Classification Models | Accuracy | Mis-Classification | F1-measure | AUC | Kappa | Recall |
|---|---|---|---|---|---|---|
| Random Forest | 0.87 | 0.13 | 0.86 | 0.93 | 0.692 | 0.83 |
| Logistic Regression | 0.78 | 0.22 | 0.778 | 0.79 | 0.493 | 0.71 |
| K-NN model (k=3) | 0.8 | 0.2 | 0.797 | 0.83 | 0.538 | 0.72 |
| K-NN model (k=10) | 0.87 | 0.13 | 0.89 | 0.84 | 0.676 | 0.93 |

Receiver operating characteristics (ROC) curves were plotted to compare the performance of the classification methods at all thresholds. These curves show the trade-off between true positive rate and false-positive rate. The highest value is attributed to the Random forest model (The area under the curve (AUC) =0.93) and the lowest area achieved in the logistic regression model. It is interesting to note that changing the number of neighbors in the KNN model has a negligible effect on the AUC value.

a) b)

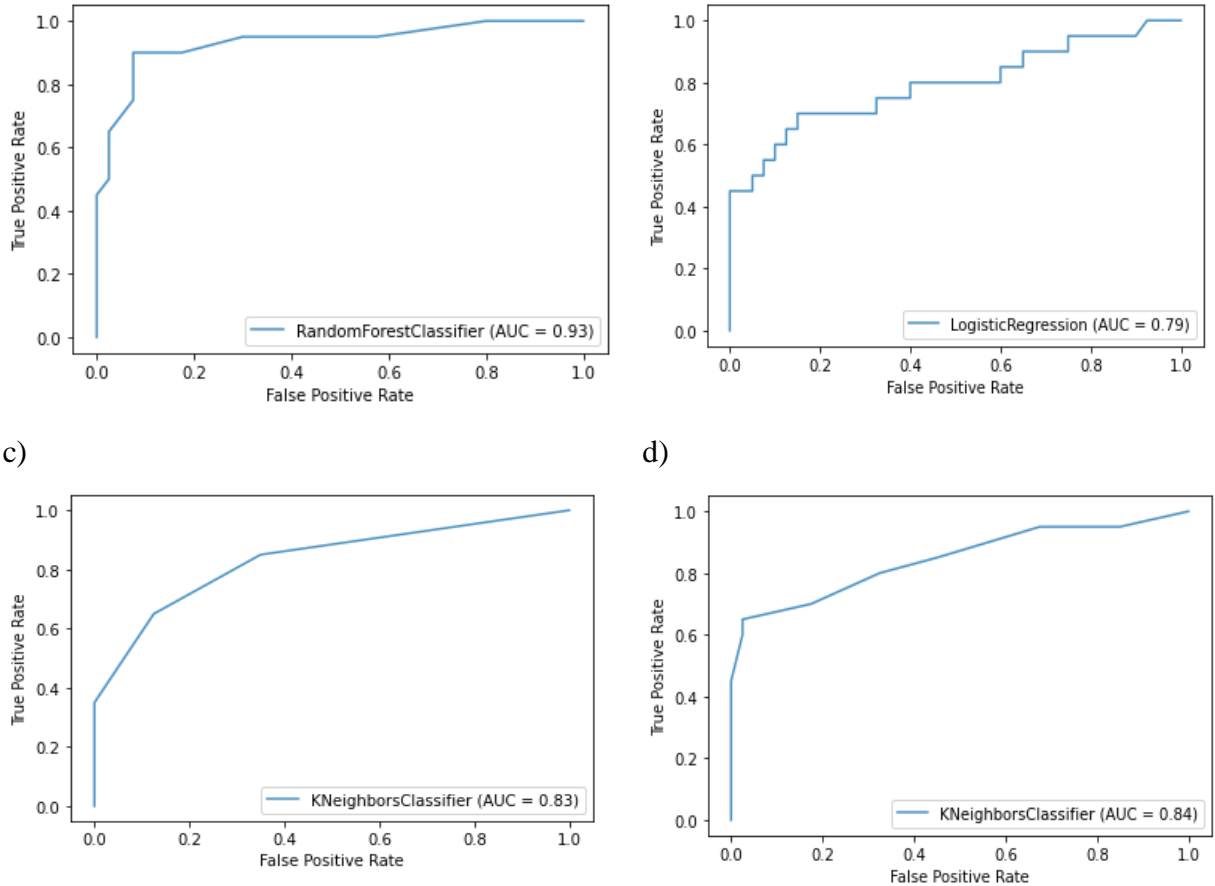

**Figure 7.** ROC plots from a) RF model, b) Logistic regression model, c) K-NN model (k=3), d) K-NN model (k=3). The highest value of AUC achieved for Random forest model and the lowest area is related to logistic regression model.

For the last approach, boosting algorithm was applied to our dataset as a class of ensemble learning that combines the prediction made by weak learners in the ensemble. To compare the performance of the decision tree with boosting algorithm, we selected AdaBoost (Adaptive Boosting) that includes short one-level decision trees added sequentially, wherein the sequence, later model tries to correct the predictions of the previous model and adjust the error.

The accuracy of the model is 0.844, with a standard deviation of 0.093. Increasing the number of trees to 10-15 trees could enhance the accuracy to 0.87, with a lower standard deviation (SD) in the case of 10 trees.

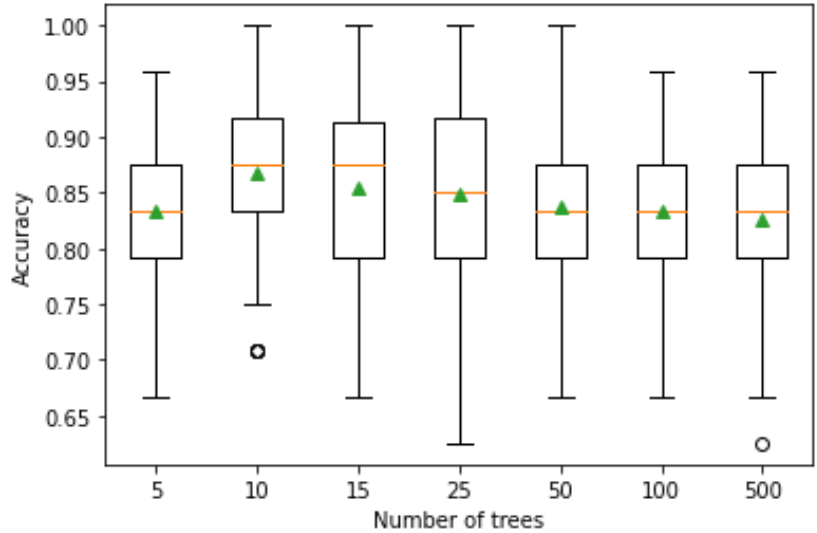

**Figure 8.** Evaluation the accuracy of AdaBoost model versus number of trees, the highest accuracy achieved by increasing the number of trees to 10 in Adaboost model.

The results of ML methods on the E-jet printing data suggested RF along with AdaBoost as the most accurate models where the effect of nozzle speed is more significant on the classification. The experimental data could successfully train the ML models that can predict the conductivity status of graphene biosensors and avoid producing a defective piece and save time and material in the E-jet process that is only one step of creating a productive biosensor.

## 4. Conclusions

In this paper, the organic graphene-based electrodes were fabricated using a high-precision E-jet printer. Three different processing parameters were changed to collect the electrical properties data that can help the supervised ML method to establish the relations between inputs and output. ML uses mathematical algorithms to classify a set of data and make predictions based on the classifications made.

The primary motivation of this work was to control the processing parameters that will eventually create defective sensors by increasing the resistance of electrodes. One of the advantages of these models is the ability to change the range of conductivity values to set the classes in real-time. Since the highest accuracy obtained from a generated decision tree was 0.71 even after decreasing the number of splits, more advanced data-driven predictive supervised learning models were used on our data.

A well-performing model is expected to control and manage the processing parameters and predict the conductivity results by performing the real-time classification models. Since printing with the e-jet technique is a time-demanding method that depends on various processing conditions, the production of nonfunctional electrodes can be stopped by performing ML prediction before spending time and material on the experiment.

## Acknowledgments

This work was supported in part by the National Science Foundation Award 2014346 and the Army Research Office Grant W911NF-18-1-0412.

## Conflict of interest

The authors declare no conflict of interest.

## References

[1]   A. Ghilan, A. P. Chiriac, L. E. Nita, A. G. Rusu, I. Neamtu, and V. M. Chiriac, "Trends in 3D Printing Processes for Biomedical Field: Opportunities and Challenges," (in English), *J Polym Environ,* vol. 28, no. 5, pp. 1345-1367, May 2020, doi: 10.1007/s10924-020-01722-x.

[2]   M. Jelbuldina, H. Younes, I. Saadat, L. Tizani, S. Sofela, and A. A. Ghaferi, "Fabrication and design of CNTs inkjet-printed based micro FET sensor for sodium chloride scale detection in oil field," *Sensors and Actuators A: Physical,* vol. 263, pp. 349-356, 2017, doi: 10.1016/j.sna.2017.06.035.

[3]   D. W. Kim, S. Y. Min, Y. Lee, and U. Jeong, "Transparent Flexible Nanoline Field-Effect Transistor Array with High Integration in a Large Area," (in English), *Acs Nano,* vol. 14, no. 1, pp. 907-918, Jan 2020, doi: 10.1021/acsnano.9b08199.

[4]   B. H. Kim *et al.*, "High-Resolution Patterns of Quantum Dots Formed by Electrohydrodynamic Jet Printing for Light-Emitting Diodes," (in English), *Nano Lett,* vol. 15, no. 2, pp. 969-973, Feb 2015, doi: 10.1021/nl503779e.

[5]   J. U. Park *et al.*, "High-resolution electrohydrodynamic jet printing," (in English), *Nat Mater,* vol. 6, no. 10, pp. 782-789, Oct 2007, doi: 10.1038/nmat1974.

[6]   A. E. N. Asli, J. S. Guo, P. L. Lai, R. Montazami, and N. N. Hashemi, "High-Yield Production of Aqueous Graphene for Electrohydrodynamic Drop-on-Demand Printing of Biocompatible Conductive Patterns," (in English), *Biosensors-Basel,* vol. 10, no. 1, Jan 2020, doi: ARTN 6

10.3390/bios10010006.

[7]   J. L. Li *et al.*, "Fabrication of three-dimensional porous scaffolds with controlled filament orientation and large pore size via an improved E-jetting technique," (in English), *J Biomed Mater Res B,* vol. 102, no. 4, pp. 651-658, May 2014, doi: 10.1002/jbm.b.33043.

[8]   Y. Wu, Z. Y. Wang, J. Y. H. Fuh, Y. S. Wong, W. Wang, and E. S. Thian, "Direct E-jet printing of three-dimensional fibrous scaffold for tendon tissue engineering," (in English), *J Biomed Mater Res B,* vol. 105, no. 3, pp. 616-627, Apr 2017, doi: 10.1002/jbm.b.33580.

[9]   M. S. Onses, E. Sutanto, P. M. Ferreira, A. G. Alleyne, and J. A. Rogers, "Mechanisms, Capabilities, and Applications of High-Resolution Electrohydrodynamic Jet Printing," *Small,* vol. 11, no. 34, pp. 4237-66, Sep 9 2015, doi: 10.1002/smll.201500593.

[10]  F. D. Prasetyo, H. T. Yudistira, V. D. Nguyen, and D. Byun, "Ag dot morphologies printed using electrohydrodynamic (EHD) jet printing based on a drop-on-demand (DOD) operation," *Journal of Micromechanics and Microengineering,* vol. 23, no. 9, 2013, doi: 10.1088/0960-1317/23/9/095028.

[11]  R. Das and S. S. Roy, "Parameter Design of High-Resolution E-Jet Micro-Fabrication Process by Taguchi Utility Approach," *Int. J. Manuf. Mater. Mech. Eng.,* vol. 8, pp. 44-58, 2018.

[12]  Y. Zhao, A. Kim, G. X. Wan, and B. C. K. Tee, "Design and applications of stretchable and self-healable conductors for soft electronics," (in English), *Nano Converg,* vol. 6, Aug 1 2019, doi: ARTN 25

10.1186/s40580-019-0195-0.

[13]  P. Cataldi *et al.*, "Carbon Nanofiber versus Graphene-Based Stretchable Capacitive Touch Sensors for Artificial Electronic Skin," (in English), *Adv Sci,* vol. 5, no. 2, Feb 2018, doi: ARTN 1700587

10.1002/advs.201700587.

[14]  B. Kang *et al.*, "Nanopatched Graphene with Molecular Self-Assembly Toward Graphene-Organic Hybrid Soft Electronics," (in English), *Adv Mater,* vol. 30, no. 25, Jun 20 2018, doi: ARTN 1706480

10.1002/adma.201706480.

[15]  K. Kwon *et al.*, "Wireless, soft electronics for rapid, multisensor measurements of hydration levels in healthy and diseased skin," (in English), *P Natl Acad Sci USA,* vol. 118, no. 5, Feb 2 2021, doi: ARTN e2020398118

10.1073/pnas.2020398118.

[16]  S. Y. Zhang, S. B. Li, Z. Z. L. Xia, and K. Y. Cai, "A review of electronic skin: soft electronics and sensors for human health," (in English), *J Mater Chem B,* vol. 8, no. 5, pp. 852-862, Feb 7 2020, doi: 10.1039/c9tb02531f.

[17]  M. Berggren and A. Richter-Dahlfors, "Organic bioelectronics," (in English), *Adv Mater,* vol. 19, no. 20, pp. 3201-3213, Oct 19 2007, doi: 10.1002/adma.200700419.

[18]  Y. Fang, X. M. Li, and Y. Fang, "Organic bioelectronics for neural interfaces," (in English), *J Mater Chem C,* vol. 3, no. 25, pp. 6424-6430, 2015, doi: 10.1039/c5tc00569h.

[19]  Y. S. Hsiao, C. W. Kuo, and P. L. Chen, "Multifunctional Graphene-PEDOT Microelectrodes for On Chip Manipulation of Human Mesenchymal Stem Cells," (in English), *Adv Funct Mater,* vol. 23, no. 37, pp. 4649-4656, Oct 4 2013, doi: 10.1002/adfm.201203631.

[20]  D. L. Gan *et al.*, "Graphene Oxide-Templated Conductive and Redox-Active Nanosheets Incorporated Hydrogels for Adhesive Bioelectronics," (in English), *Adv Funct Mater,* vol. 30, no. 5, Jan 2020, doi: ARTN 1907678

[21] L. Wang *et al.*, "Inkjet jet failures detection and droplets speed monitoring using piezo self-sensing," *Sensors and Actuators A: Physical,* vol. 313, 2020, doi: 10.1016/j.sna.2020.112178.

[22] G. D. Goh, S. L. Sing, and W. Y. Yeong, "A review on machine learning in 3D printing: applications, potential, and challenges," (in English), *Artif Intell Rev,* vol. 54, no. 1, pp. 63-94, Jan 2021, doi: 10.1007/s10462-020-09876-9.

[23] N. Hashemi and N. N. Clark, "Artificial neural network as a predictive tool for emissions from heavy-duty diesel vehicles in Southern California," *International Journal of Engine Research,* vol. 8, no. 4, pp. 321-336, 2007, doi: 10.1243/14680874jer00807.

[24] T. Leng *et al.*, "Screen-Printed Graphite Nanoplate Conductive Ink for Machine Learning Enabled Wireless Radiofrequency-Identification Sensors," (in English), *Acs Appl Nano Mater,* vol. 2, no. 10, pp. 6197-6208, Oct 2019, doi: 10.1021/acsanm.9b01034.

[25] A. Caggiano, J. J. Zhang, V. Alfieri, F. Caiazzo, R. Gao, and R. Teti, "Machine learning-based image processing for on-line defect recognition in additive manufacturing," (in English), *Cirp Ann-Manuf Techn,* vol. 68, no. 1, pp. 451-454, 2019, doi: 10.1016/j.cirp.2019.03.021.

[26] H. Zhang and S. K. Moon, "Reviews on Machine Learning Approaches for Process Optimization in Noncontact Direct Ink Writing," *ACS Appl Mater Interfaces,* May 27 2021, doi: 10.1021/acsami.1c04544.

[27] "<1-s2.0-S2351978921000524-main.pdf>."

[28] Q. Zhao, H. Yang, J. Liu, H. Zhou, H. Wang, and W. Yang, "Machine learning-assisted discovery of strong and conductive Cu alloys: Data mining from discarded experiments and physical features," *Materials & Design,* vol. 197, 2021, doi: 10.1016/j.matdes.2020.109248.

[29] J. Shi *et al.*, "Learning-Based Cell Injection Control for Precise Drop-on-Demand Cell Printing," (in English), *Ann Biomed Eng,* vol. 46, no. 9, pp. 1267-1279, Sep 2018, doi: 10.1007/s10439-018-2054-2.

[30] J. Shi, J. Song, B. Song, and W. F. Lu, "Multi-Objective Optimization Design through Machine Learning for Drop-on-Demand Bioprinting," *Engineering,* vol. 5, no. 3, pp. 586-593, 2019, doi: 10.1016/j.eng.2018.12.009.

[31] D. Wu and C. Xu, "Predictive Modeling of Droplet Formation Processes in Inkjet-Based Bioprinting," *Journal of Manufacturing Science and Engineering,* vol. 140, no. 10, 2018, doi: 10.1115/1.4040619.

[32] M. Akbari, L. Sydanheimo, J. Juuti, J. Vuorinen, and L. Ukkonen, "Characterization of Graphene-Based Inkjet Printed Samples on Flexible Substrate for Wireless Sensing Applications," (in English), *2014 Ieee Rfid Technology and Applications Conference (Rfid-Ta),* pp. 135-139, 2014. [Online]. Available: <Go to ISI>://WOS:000355257100026.

[33] S. F. Pei and H. M. Cheng, "The reduction of graphene oxide," (in English), *Carbon,* vol. 50, no. 9, pp. 3210-3228, Aug 2012, doi: 10.1016/j.carbon.2011.11.010.

[34] R. K. L. Tan *et al.*, "Graphene as a flexible electrode: review of fabrication approaches," *Journal of Materials Chemistry A,* vol. 5, no. 34, pp. 17777-17803, 2017, doi: 10.1039/c7ta05759h.

[35] F. Torrisi *et al.*, "Inkjet-Printed Graphene Electronics," (in English), *Acs Nano,* vol. 6, no. 4, pp. 2992-3006, Apr 2012, doi: 10.1021/nn2044609.

[36] L. Zhang, "Characteristics of drop-on-demand droplet jetting with effect of altered geometry of printhead nozzle," *Sensors and Actuators A: Physical,* vol. 298, 2019, doi: 10.1016/j.sna.2019.111591.

[37] M. Lotya *et al.*, "Liquid Phase Production of Graphene by Exfoliation of Graphite in Surfactant/Water Solutions," (in English), *J Am Chem Soc,* vol. 131, no. 10, pp. 3611-3620, Mar 18 2009, doi: 10.1021/ja807449u.

[38] K. R. Paton *et al.*, "Scalable production of large quantities of defect-free few-layer graphene by shear exfoliation in liquids," (in English), *Nat Mater,* vol. 13, no. 6, pp. 624-630, Jun 2014, doi: 10.1038/Nmat3944.

[39] S. Ahadian *et al.*, "Facile and green production of aqueous graphene dispersions for biomedical applications," (in English), *Nanoscale,* vol. 7, no. 15, pp. 6436-6443, 2015, doi: 10.1039/c4nr07569b.